\documentclass[11pt]{article}
\usepackage[T1]{fontenc}
\usepackage[utf8]{inputenc}
\usepackage{lmodern}
\usepackage{microtype}
\usepackage{graphicx,booktabs,tabularx,array}
\usepackage{colortbl}
\usepackage{flafter}
\usepackage[numbers,sort]{natbib}
\usepackage[colorlinks=true,linkcolor=black,citecolor=black,urlcolor=black]{hyperref}
\usepackage{nanostyle}
\usepackage{nanopage}
\usepackage{caption}
\usepackage{float}
\usepackage{wrapfig}
\usepackage{tikz}
\usetikzlibrary{arrows.meta, calc, decorations.pathreplacing}
\crefname{figure}{Figure}{Figures}
\crefname{table}{Table}{Tables}
\crefname{section}{Section}{Sections}
\crefname{subsection}{Section}{Sections}
\crefname{appendix}{Appendix}{Appendices}
\crefname{subappendix}{Appendix}{Appendices}
\crefname{equation}{Eq.}{Eqs.}

\newcolumntype{Y}{>{\raggedright\arraybackslash}X}
\newcommand{\FTT}{FTT}
\newcommand{\RMSNorm}{N}

\newcommand{\Att}{\operatorname{Att}}
\newcommand{\todo}[1]{\textcolor{red!70!black}{\textbf{TODO}\if\relax\detokenize{#1}\relax\else: #1\fi}}
\title{Field-Time Transformers: Universal Model Architecture for Imperfect-Information Games by Structured Attention}
\author{
    Zhiyuan Fan\\
    MIT\\
    \texttt{fanzy@mit.edu}
    \And
    Gabriele Farina \\
    MIT \\
    \texttt{gfarina@mit.edu}
}
\date{\today}
\hypersetup{pdftitle={Field-Time Transformers: Universal Model Architecture for Imperfect-Information Games by Structured Attention},pdfauthor={Zhiyuan Fan, Gabriele Farina},pdfsubject={Preprint}}
\graphicspath{{figures/}}
\begin{document}
\maketitle
\begin{abstract}
Self-play reinforcement learning in imperfect-information games is an active research area. While most of the recent progress focuses on the learning algorithm side, there is surprisingly little focus on the model architecture. Most agents rely on recurrent methods such as LSTM to encode the history information, which tends to compress the entire game into one state. However, this is different from how people play unfamiliar games in the real world: a useful reasoning strategy is to track the individual components of the game and analyze their self-interaction or cross-interaction over time. For example, a player can inspect the card colors or ranks in Hanabi independently rather than mixing everything.

In this work, we propose the Field-Time Transformer (FTT) inspired by this thinking pattern. FTT takes the history of different components, or \emph{fields}, as inputs, and alternately uses field attention among the fields at the same timestep and temporal attention along each field's history. As almost all games admit such a decomposable structure, this design provides a general architecture for imperfect-information games.

FTT presents strong empirical performance on a variety of games. For cooperative games where an absolute score is available, we show that the FTT agent is more sample efficient compared to the previous LSTM baselines. It reaches the state-of-the-art scores of 24.78, 24.96, 24.95, and 24.86, with perfect rates of 84.75\%, 97.12\%, 95.84\%, and 88.80\% in two- to five-player Hanabi, respectively. In more competitive game environments, Dou Dizhu, Guan Dan, and Dark Chess, the FTT agents also achieve stronger performance compared to the released agents in a variety of games, including those with test-time improvement.
\end{abstract}

\section{Introduction}
\label{sec:introduction}

An increasingly popular approach for solving imperfect-information games is self-play reinforcement learning, as recent training recipes have demonstrated its strong performance. It has produced the best-known agents in a variety of strategic games and even achieved superhuman performance; examples include Stratego~\citep{perolat2022mastering,sokota2026ataraxos}, Hanabi~\citep{hu2019sad,hu2021offbelief}, Mahjong~\citep{li2020suphx,fu2022ach,tencent2023luckyj}, Dou Dizhu~\citep{zha2021douzero,yang2022perfectdou,fan2026vrpo}, and Generals.io~\citep{straka2026generals}. Previous studies generally focus on the training algorithm. Methods including regularized policy gradients~\citep{sokota2022unified}, centralized critics~\citep{yu2022mappo}, and lower-variance advantage estimates~\citep{fan2026vrpo} improve the convergence speed as well as the stability during training.

However, there has been surprisingly little attention to the model architecture that encodes the policy. Since the policy of an imperfect-information game might rely on small changes in the game history, the policy generally needs to account for any factor of the past. The most common approach is thus to compress the game history into a single hidden state with recurrent neural networks, with the network taking a flat hand-designed feature vector as input~\citep{bard2019hanabi}. This approach is not ideal as the long-term dependency is not naturally encoded in a small recurrent state and is generally hard to distinguish from noise. More recent approaches choose to replace the recurrent network over states with a Transformer over a stream of events~\citep{fan2026vrpo}, or manually select the long-term features and parse the game states with a Transformer on the current time frame~\citep{sokota2026ataraxos,straka2026generals}.

People generally follow a more structured thinking pattern different from these agents. A more natural reasoning strategy is to track the persistent components in the games. In card games like Hanabi, a natural way is to think over the cards grouped by different colors individually, dividing the task into smaller groups and trying to come up with a policy that focuses on only a small portion of the cards, then integrating the thoughts from different parts to synthesize a reasoning move. Compared to the straightforward approach that investigates the game as a complex object as a whole, corresponding to compressing the entire game history into a state vector, this divide-and-conquer reasoning pattern can have better alignment with the structure of the game and thus can enable more effective reasoning.

In this work, we introduce the Field-Time Transformer (\FTT) inspired by this structured reasoning pattern. \FTT\ keeps track of the persistent components, referred to as \emph{fields}, and treats them as first-class elements of the model architecture. Rather than compressing the game history into a single vector stream as in the recurrent neural network approaches, \FTT\ maintains a set of parallel streams associated with the fields over timesteps. The field streams are processed by attention mechanisms applied in a factorized and interleaved manner, across fields at each timestep and across timesteps within each field, similar to the divided space-time attention for video modeling in ViViT~\citep{arnab2021vivit} and TimeSformer~\citep{bertasius2021timesformer}. The action heads are further organized according to the game's structure such that each action is read out from the field it relates to. This creates a structured representation of the action space rather than using a flat output layer with thousands of largely unrelated heads.

We benchmark the performance of \FTT\ on a variety of games and present strong performance across all of them. In the cooperative game Hanabi, the \FTT\ agent reaches self-play scores of 24.78, 24.96, 24.95, and 24.86 in two- to five-player games. These scores exceed every published result we are aware of, including those with test-time policy improvement. Under the same amount of environment timesteps and matched actor computation, the \FTT\ agent learns much faster and reaches higher final scores and perfect-game rates compared to the LSTM actors. In the competitive games, the \FTT\ agent outperforms the previous state-of-the-art bot Ataraxos~\citep{sokota2026ataraxos} in Dou Dizhu by 0.19 points per deal with its test-time policy improvement enabled, and its base policy by 0.24. In Guan Dan, the \FTT\ agent wins 72.7\% of the hands against the public baseline DanZero~\citep{lu2023danzero}. In Dark Chess, the \FTT\ agent scores 59.6\% against our reimplementation of the search of the superhuman Obscuro~\citep{zhang2026obscuro}, which solves subgames at test time, with a search tree of one million nodes.

To summarize, our main contributions are:
\begin{itemize}[leftmargin=*,itemsep=0pt,topsep=0pt]
\item We propose \FTT, an architecture for imperfect-information game agents that models the observation history as parallel field streams evolving over time.
\item Experiments show that \FTT\ agents achieve state-of-the-art performance in Hanabi, Dou Dizhu, and Guan Dan, and score above one half against our reimplementation of the search of Obscuro in Dark Chess.
\end{itemize}
\section{Related Work}
\label{sec:related-work}

\paragraph{Model Architecture for Imperfect-Information Games.}
In imperfect-information games, different game histories imply different world states, and thus agents need to make decisions based on the information in the past. A standard approach to processing temporal information is to encode each timestep into a frame and process it using an LSTM~\citep{hochreiter1997lstm}. In Hanabi, the feature vector is created using a flat, hand-designed encoding on the current observation~\citep{bard2019hanabi,hu2019sad,hu2020otherplay,hu2021offbelief,sokota2026ataraxos}. The superhuman video game agents AlphaStar~\citep{vinyals2019grandmaster} and OpenAI Five~\citep{berner2019dota} further use self-attention or pooling to combine the units to form the observation. Another line of work uses hand-crafted observation tensors to include past move history. In Dou Dizhu and Guan Dan, DouZero~\citep{zha2021douzero} and PerfectDou~\citep{yang2022perfectdou} encode the hands, played cards, and the most recent 15 moves into card matrices processed with a CNN. The Guan Dan agent DanZero~\citep{lu2023danzero} uses a similar encoding but uses a fully connected network as the main trunk. In Stratego, DeepNash~\citep{perolat2022mastering} stacks the last 40 moves into a board tensor and processes it with a CNN. More recent approaches, including Ataraxos in Stratego~\citep{sokota2026ataraxos} and AverageJoe in Generals.io~\citep{straka2026generals}, use hand-designed history information and process the board with a Transformer.

\paragraph{Factorized Attention.}
The tasks of video models generally need to comprehend patches across the frames. While a full computation among all patches of the video can be inefficient, one of the most standard approaches is to factorize the computation into the spatial axis and the temporal axis, as P3D~\citep{qiu2017pseudo3d}, R(2+1)D~\citep{tran2018closer} and S3D~\citep{xie2018rethinking} do for 3D convolutions. TimeSformer~\citep{bertasius2021timesformer} brings this factorization to attention: each block first attends to the same patch location across frames and then to the other patches of the same frame.
ViViT~\citep{arnab2021vivit} compares several such factorizations with joint attention over all space-time tokens, and other models apply the two attentions separately or in stages~\citep{zhang2021vidtr,neimark2021vtn,girdhar2021avt}. Another line of work keeps joint attention but restricts it to local windows~\citep{liu2022videoswin}, pooled keys and values~\citep{fan2021mvit} or motion trajectories~\citep{patrick2021motionformer}. Video generation models mostly use the factorized form~\citep{ho2022vdm,singer2023makeavideo,blattmann2023align}, and Latte~\citep{ma2025latte} finds alternating spatial and temporal blocks the best of four decompositions. In most of these models the temporal attention covers a whole clip in both directions, while Phenaki~\citep{villegas2023phenaki} and the world model Genie~\citep{bruce2024genie} make it causal. Outside video, Scene Transformer~\citep{ngiam2022scene} alternates attention across time and across the agents in driving scenes.

\section{The Field-Time Transformer}
\label{sec:method}

\begin{wrapfigure}{r}{0.42\linewidth}
\vspace{-\baselineskip}
\centering
\begin{tikzpicture}[x=1cm, y=1cm, font=\fontsize{9}{10.8}\selectfont,
    >={Stealth[length=3pt, width=2.5pt]},
    layer/.style={draw=black!80, line width=0.5pt, rounded corners=2pt,
        minimum width=3.6cm, minimum height=0.44cm, inner xsep=1.5pt,
        inner ysep=0.8pt, align=center},
    norm/.style={layer, minimum height=0.28cm, inner ysep=0.7pt,
        fill=fancyAmber!28, font=\fontsize{8}{9.6}\selectfont},
    side/.style={layer, minimum width=2.3cm, minimum height=0.38cm,
        font=\fontsize{8}{9.6}\selectfont},
    panel/.style={draw=black!60, fill=black!3, rounded corners=4pt, line width=0.5pt},
    flow/.style={->, line width=0.6pt, draw=black!80},
    res/.style={->, line width=0.6pt, draw=black!80, rounded corners=3pt},
    plus/.style={circle, draw=black!80, line width=0.5pt, inner sep=0pt,
        minimum size=0.22cm, font=\fontsize{7}{8}\selectfont, fill=white},
    lbl/.style={inner sep=1pt},
]
\newcommand{\sttsublayer}[4]{%
  \coordinate (b#1) at (0, {#2+0.02});
  \node[norm] (n#1) at (0, {#2+0.25}) {RMSNorm};
  \node[layer, fill=#3] (o#1) at (0, {#2+0.70}) {#4};
  \node[plus] (p#1) at (0, {#2+1.14}) {$+$};
  \draw[flow] (n#1.north) -- (o#1.south);
  \draw[flow] (o#1.north) -- (p#1.south);
  \draw[res] (b#1) -- (-2.15, {#2+0.02}) -- (-2.15, {#2+1.14}) -- (p#1.west);
  \fill[black!80] (b#1) circle (0.025);
}
\newcommand{\sttinput}[2]{%
  \node[lbl] (#1obs) at (0, 0.08) {#2};
  \node[layer, fill=fancyColor!25] (#1tok) at (0, 0.59) {Token Embedding};
  \draw[flow] (#1obs) -- (#1tok.south);
  \node[plus] (#1fsum) at (0, 1.16) {$+$};
  \draw[flow] (#1tok.north) -- (#1fsum.south);
  \node[side, fill=fancyColor!25] (#1femb) at (1.85, 1.16) {Field Embedding};
  \draw[flow] (#1femb.west) -- (#1fsum.east);
}
\sttinput{a}{Observation $\vo_t$}
\begin{scope}[yshift=-0.62cm]
\draw[panel] (-2.50, 2.12) rectangle (2.20, 6.28);
\sttsublayer{at}{2.24}{fancyBlue!25}{Temporal Attention}
\sttsublayer{as}{3.58}{fancyTeal!28}{Field Attention}
\sttsublayer{af}{4.92}{black!9}{Feed Forward}
\draw[flow] (pat.north) -- (nas.south);
\draw[flow] (pas.north) -- (naf.south);
\node[lbl, anchor=east] at (-2.65, 4.20) {$L\,\times$};
\node[norm] (anorm) at (0, 6.54) {RMSNorm};
\node[layer, fill=fancyRed!22] (alin) at (0, 7.02) {Linear};
\node[layer, fill=fancyRed!22] (asoft) at (0, 7.60) {Softmax};
\node[lbl] (api) at (0, 8.11) {Action Probabilities $\pi(\,\cdot\mid \vo_{1:t})$};
\draw[flow] (paf.north) -- (anorm.south);
\draw[flow] (anorm.north) -- (alin.south);
\draw[flow] (alin.north) -- (asoft.south);
\draw[flow] (asoft.north) -- (api);
\end{scope}
\draw[flow] (afsum.north) -- (nat.south);

\end{tikzpicture}
\caption{\FTT. Each arrow carries one vector per field, and the block is repeated $L$ times.}
\label{fig:architecture}
\vspace{-\baselineskip}
\end{wrapfigure}
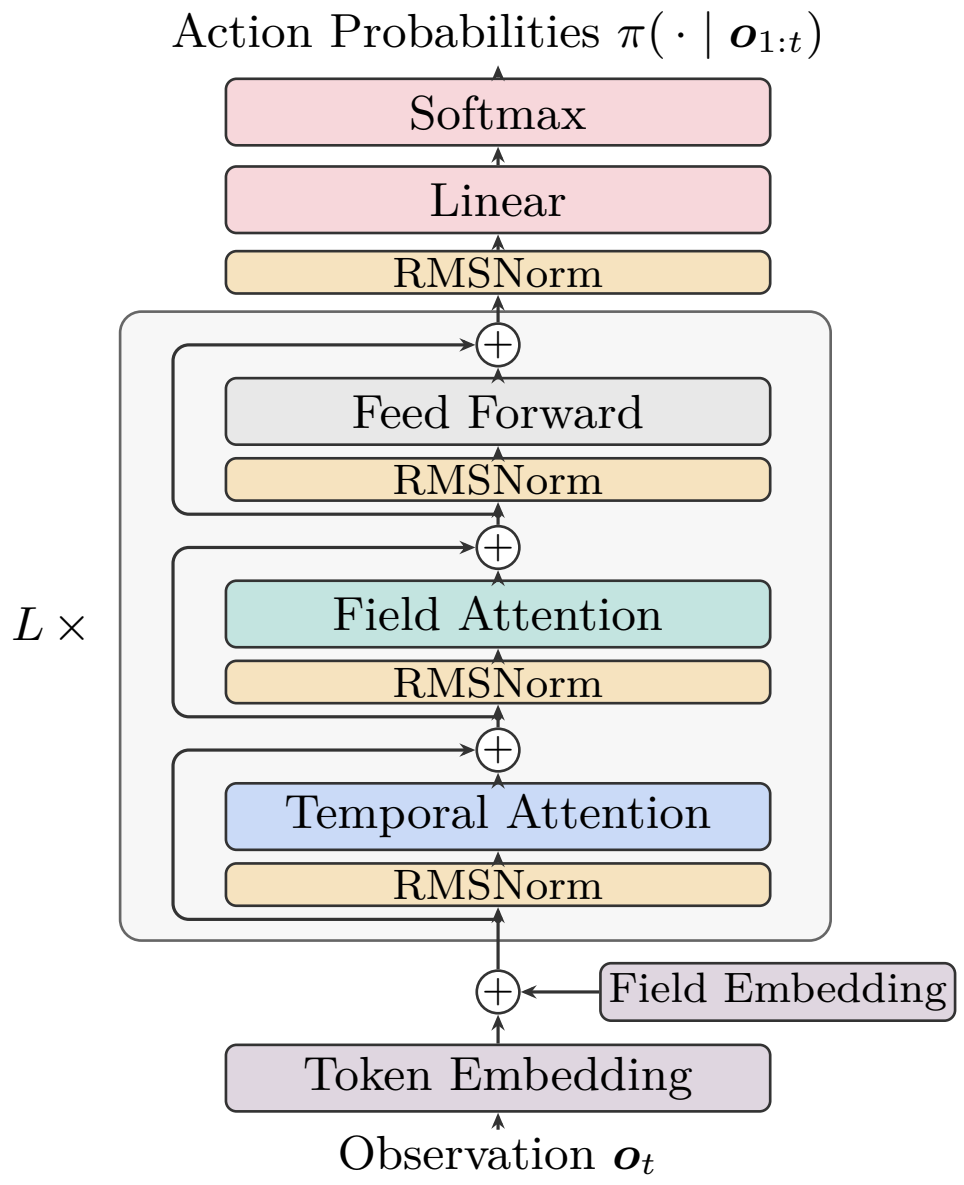

In this section, we introduce the Field-Time Transformer (\FTT), shown in \cref{fig:architecture}. We first describe how to represent the game using fields in \cref{sec:fields}. We then explain how the field streams are processed over time and how the actions are read out in \cref{sec:backbone}. Further details on individual games are provided in Appendix~\ref{app:interfaces}.

\subsection{Representing Games with Fields}

\label{sec:fields}

The intuition behind \FTT\ is to track persistent components of a game throughout an episode, which we refer to as \emph{fields}. Fields can be chosen flexibly according to the structure of each game, preferably capturing strategically relevant aspects. At every decision step a field holds what the player currently knows about its component, and over an episode this forms a data stream that records all events relevant to it. Information that does not naturally belong to any specific field is instead assigned to an additional \emph{global field}. The information corresponding to the field is generally sparse with respect to the possible features. As a result, we prefer to use a fixed list of slots, each holding one token from a large feature vocabulary. While this is equivalent to a sparse 0-1 encoding, it allows the model to have an encoding that supports a variety of tokens.

Take Hanabi as an example: the color and the rank are two persistent components of the game, so we set up one field for each color and one for each rank. Each field then records the history of its component: for each color field, we record the height of that color's firework, which ranks have been played or discarded, which of the partner's cards are that color, and which cards in any player's hand could still be that color. The field of a rank records similar information but corresponding to the ranks. Additional information like players' seats, who is acting, and the number of hint and life tokens goes to the global field stream. Other games like Dou Dizhu take the rank as the fields, and each field counts the cards of that rank in the player's hand and played by each player. Detailed information of other games can be found in Appendix~\ref{app:interfaces}.

We set up the action readout according to the field each action relates to. An action that concerns the component of a local field is read out from that field; otherwise it is attached to the global field. In Hanabi, we read out hints that name a color or a rank from the local field corresponding to that color or rank. Playing or discarding a card from the hand names no color or rank, and we thus attach it to the global field. In Dou Dizhu, actions that name a rank are read out from that rank's field, while pass, bidding and doubling go to the global field. In games with turns, a player that is not to act still observes the step and takes a \textsc{Noop}, which the global field reads out. \Cref{tab:games} lists the number of fields, token slots and outputs of every game.

\begin{table}[!htb]
\centering\small
\setlength{\tabcolsep}{3pt}
\begin{tabular}{@{}llcccccc@{}}
\toprule
 & & & \multicolumn{2}{c}{Token Slots} & & \multicolumn{2}{c}{Action Readouts}\\
\cmidrule(lr){4-5}\cmidrule(l){7-8}
Game & Field Type & Field Count $K$ & Local & Global & Total Actions & Local & Global\\
\midrule
Hanabi 2P & color or rank & 10 & 26 & 9 & 21 & 1 & 11\\
Hanabi 3P & color or rank & 10 & 36 & 10 & 31 & 2 & 11\\
Hanabi 4P & color or rank & 10 & 39 & 11 & 39 & 3 & 9\\
Hanabi 5P & color or rank & 10 & 47 & 12 & 49 & 4 & 9\\
Dou Dizhu & card rank & 15 & 6 & 30 & 134 & 10 & 11\\
Guan Dan & card rank & 15 & 24 & 34 & 172 & 13 & 4\\
Dark Chess & board square & 64 & 1 & 6 & 4,865 & 76 & 1\\
\bottomrule
\end{tabular}

\caption{The parameters of the games considered in this work. Action readout gives the output width per field. Unused rank outputs in the card games are omitted from the action set, and the actions include \textsc{Noop} where applicable.}
\label{tab:games}
\end{table}

\subsection{Field Attention and Temporal Attention}
\label{sec:backbone}

\paragraph{Input Embedding.}
Denote by $\vo_t$ the observation tensor at decision step $t$ such that $\vo_t[k,j]$ contains the token in slot $j$ of field $k$. The embedding layer of \FTT\ embeds the tokens into a vector with a shared table $\mE$ together with the field's own embedding $\vp_k$:
\begin{equation}
\vx^{0}_{t,k}=\vp_k+\sum_{j}\mE\bigl[\vo_t[k,j]\bigr],\qquad k \in \llbracket 0, K \rrbracket.
\label{eq:embedding}
\end{equation}
Here $k=0,\dots,K-1$ are the local fields and $k=K$ is the global field, which takes its tokens from a table of its own. While $\vp_k$ is generally a field-specific encoding, for structured fields like a board, $\vp_k$ can be chosen as the sum of its row and column embeddings.

\paragraph{The \FTT\ Block.}
The backbone of the network is a stack of $L$ identical blocks, each containing three sub-layers: a temporal attention sub-layer, a field attention sub-layer, and a feed-forward network:
\begin{align}
    \vu_{t,k}&=\vx^{\ell-1}_{t,k}+\Att^{T}_{\ell}\bigl(\RMSNorm\bigl(\vx^{\ell-1}_{t,k}\bigr);\;\bigl\{\RMSNorm\bigl(\vx^{\ell-1}_{\tau,k}\bigr)\bigr\}_{\tau\in\cW_t}\bigr),\label{eq:temporal}\\
    \vv_{t,k}&=\vu_{t,k}+\Att^{F}_{\ell}\bigl(\RMSNorm\bigl(\vu_{t,k}\bigr);\;\bigl\{\RMSNorm\bigl(\vu_{t,k'}\bigr)\bigr\}_{k'=0}^{K}\bigr),\label{eq:spatial}\\
    \vx^{\ell}_{t,k}&=\vv_{t,k}+F_{\ell}\bigl(\RMSNorm\bigl(\vv_{t,k}\bigr)\bigr),\label{eq:ffn}
\end{align}
where $\RMSNorm$ is the RMS normalization~\citep{zhang2019rmsnorm} applied to the input of each sub-layer, $\Att(\vq;\,\vc_1,\dots,\vc_n)$ is multi-head attention of a query over a set of context vectors~\citep{vaswani2017attention} and $F_\ell$ is a gated feed-forward network~\citep{shazeer2020glu} applied to each field separately.

In the \emph{temporal attention}~\eqref{eq:temporal}, we adopt sliding-window attention~\citep{beltagy2020longformer} with RoPE~\citep{su2021roformer} so the attention is distributed over the last $W$ timesteps of the episode of the same field. The \emph{field attention}~\eqref{eq:spatial} then takes all fields at the timestep and applies a self-attention across all the fields of the current time step. The feed-forward network uses the SwiGLU activation.

This architecture can be considered as a counterpart of the divided space-time attention of TimeSformer~\citep{bertasius2021timesformer} with fields in place of image patches. The factorization keeps the attention affordable compared with a full attention over all fields and timesteps, and since the fields follow the structure of the game, the separation is also aligned with the common reasoning pattern of people.

We further remark that the temporal attention is causal and windowed, which allows fast inference: KV caches are maintained for each temporal attention sub-layer, so that when a new observation $\vo_t$ arrives, only the states of step $t$ need to be computed. This makes \FTT\ comparable in efficiency to the previous LSTM approaches.

\paragraph{Action Readout.}
\label{sec:readout}
Denote by $\cA_t$ the legal actions at timestep $t$.
Every action $a$ is assigned to a field $k(a)$, either local or global, and to an output $s(a)$ of the readout, which has a weight vector $\vw_{s}$ and a bias $b_{s}$ for each output $s$. The action reads its field's final representation $\vh_{t,k}=\RMSNorm(\vx^{L}_{t,k})$ through its output:
\begin{equation}
    z_t(a)=\vw_{s(a)}^{\top}\vh_{t,k(a)}+b_{s(a)},
    \qquad
    \pi(a\mid \vo_{1:t})=\frac{\exp z_t(a)}{\sum_{a'\in\cA_t}\exp z_t(a')}\quad\text{for } a\in\cA_t.
    \label{eq:head}
\end{equation}
All local fields share the same outputs, so two actions of the same kind on different fields, such as hints to the same partner that name different colors, use the same weights and differ only in the field they read. The global field has outputs of its own, including the one of \textsc{Noop}. For structured games that contain a board, such as Dark Chess, a move is read out from the field of its source square, with one output for each displacement, knight move and promotion.

\section{Experiments}
\label{sec:experiments}

In this section, we evaluate the performance of \FTT\ in a variety of games. The models are trained using VRPO~\citep{fan2026vrpo}, with the hyperparameters listed in Appendix~\ref{app:training}. The detailed description of each game can be found in Appendix~\ref{app:games}.

\subsection{Hanabi: A Case Study on Different Architectures}
\label{sec:hanabi}

\begin{table}[!t]
\centering\small
\setlength{\tabcolsep}{4pt}
\begin{tabular}{@{}lclr@{}lr@{}l@{}}
\toprule
Game & \shortstack{Test-time policy\\improvement} & Agent & \multicolumn{2}{c}{Score} & \multicolumn{2}{c}{Perfect rate (\%)}\\
\midrule
Hanabi 2P & No & \textbf{FTT} & $\mathbf{24.777}$ & ${}\pm\mathbf{0.005}$ & $\mathbf{84.747}$ & ${}\pm\mathbf{0.280}$\\
 &  & SAD~\citep{hu2019sad} & $24.010$ & ${}\pm0.010$ & $52.390$ & \\
 &  & SAD + Aux~\citep{hu2019sad} & $24.080$ & ${}\pm0.010$ & $56.090$ & \\
 &  & Ataraxos~\citep{sokota2026ataraxos} & $24.454$ & ${}\pm0.008$ & $69.580$ & ${}\pm0.330$\\
 &  & Forkel et al.~\citep{forkel2025entropy} & $24.510$ & ${}\pm0.020$ &  & \\
\arrayrulecolor{black!25}\cmidrule(l){2-7}\arrayrulecolor{black}
 & \emph{Yes} & SPARTA~\citep{lerer2019sparta} & $24.610$ & ${}\pm0.010$ & $75.500$ & \\
 &  & Ataraxos~\citep{sokota2026ataraxos} & $24.654$ & ${}\pm0.007$ & $77.530$ & ${}\pm0.420$\\
\midrule
Hanabi 3P & No & \textbf{FTT} & $\mathbf{24.962}$ & ${}\pm\mathbf{0.002}$ & $\mathbf{97.120}$ & ${}\pm\mathbf{0.111}$\\
 &  & SAD~\citep{hu2019sad} & $23.930$ & ${}\pm0.010$ & $48.050$ & \\
 &  & SAD + Aux~\citep{hu2019sad} & $23.810$ & ${}\pm0.010$ & $49.740$ & \\
 &  & Ataraxos~\citep{sokota2026ataraxos} & $24.612$ & ${}\pm0.008$ & $77.610$ & ${}\pm0.290$\\
 &  & Forkel et al.~\citep{forkel2025entropy} & $24.760$ & ${}\pm0.020$ &  & \\
\arrayrulecolor{black!25}\cmidrule(l){2-7}\arrayrulecolor{black}
 & \emph{Yes} & SPARTA~\citep{lerer2019sparta} & $24.610$ & ${}\pm0.010$ & $74.500$ & \\
 &  & Ataraxos~\citep{sokota2026ataraxos} & $24.863$ & ${}\pm0.005$ & $89.900$ & ${}\pm0.300$\\
\midrule
Hanabi 4P & No & \textbf{FTT} & $\mathbf{24.949}$ & ${}\pm\mathbf{0.001}$ & $\mathbf{95.840}$ & ${}\pm\mathbf{0.083}$\\
 &  & SAD~\citep{hu2019sad} & $23.810$ & ${}\pm0.010$ & $41.450$ & \\
 &  & SAD + Aux~\citep{hu2019sad} & $23.470$ & ${}\pm0.010$ & $33.870$ & \\
 &  & Ataraxos~\citep{sokota2026ataraxos} & $24.466$ & ${}\pm0.008$ & $68.520$ & ${}\pm0.330$\\
 &  & Forkel et al.~\citep{forkel2025entropy} & $24.590$ & ${}\pm0.040$ &  & \\
\arrayrulecolor{black!25}\cmidrule(l){2-7}\arrayrulecolor{black}
 & \emph{Yes} & SPARTA~\citep{lerer2019sparta} & $24.490$ & ${}\pm0.010$ & $64.200$ & \\
 &  & Ataraxos~\citep{sokota2026ataraxos} & $24.852$ & ${}\pm0.005$ & $88.400$ & ${}\pm0.320$\\
\midrule
Hanabi 5P & No & \textbf{FTT} & $\mathbf{24.864}$ & ${}\pm\mathbf{0.004}$ & $\mathbf{88.800}$ & ${}\pm\mathbf{0.342}$\\
 &  & SAD~\citep{hu2019sad} & $23.010$ & ${}\pm0.010$ & $13.930$ & \\
 &  & SAD + Aux~\citep{hu2019sad} & $22.250$ & ${}\pm0.010$ & $7.330$ & \\
 &  & Ataraxos~\citep{sokota2026ataraxos} & $23.485$ & ${}\pm0.012$ & $26.810$ & ${}\pm0.310$\\
 &  & Forkel et al.~\citep{forkel2025entropy} & $24.090$ & ${}\pm0.060$ &  & \\
\arrayrulecolor{black!25}\cmidrule(l){2-7}\arrayrulecolor{black}
 & \emph{Yes} & SPARTA~\citep{lerer2019sparta} & $23.910$ & ${}\pm0.010$ & $32.900$ & \\
 &  & Ataraxos~\citep{sokota2026ataraxos} & $24.410$ & ${}\pm0.009$ & $58.050$ & ${}\pm0.490$\\
\bottomrule
\end{tabular}

\caption{Hanabi self-play score and perfect rate. Methods with test-time policy improvement serve as performance references. \FTT's SEM is across training seeds, as is that of \citet{forkel2025entropy}, whereas the other published errors are over evaluation games; unreported values are left blank. The reported uncertainty is $1.0 \times \mathrm{SEM}$.}
\label{tab:hanabi-published}
\end{table}

We first benchmark the performance of \FTT\ in Hanabi~\citep{bard2019hanabi}. Hanabi is a cooperative card game in which players are not allowed to see their own hand of cards but only their teammates'. The goal of the players is to play a total of $25$ cards in order. The score is the number of cards played, and a game in which all $25$ cards are played is called \emph{perfect}.

In Hanabi, the local fields of \FTT\ are the five colors and the five ranks, encoded as in \cref{sec:fields}. Each hint is read out from the field of the color or rank it names, with one output per partner, while playing or discarding each card of the hand and the \textsc{Noop} of a player who is not acting are read out from the global field.

\begin{figure}[t]
\centering
\includegraphics[width=\linewidth]{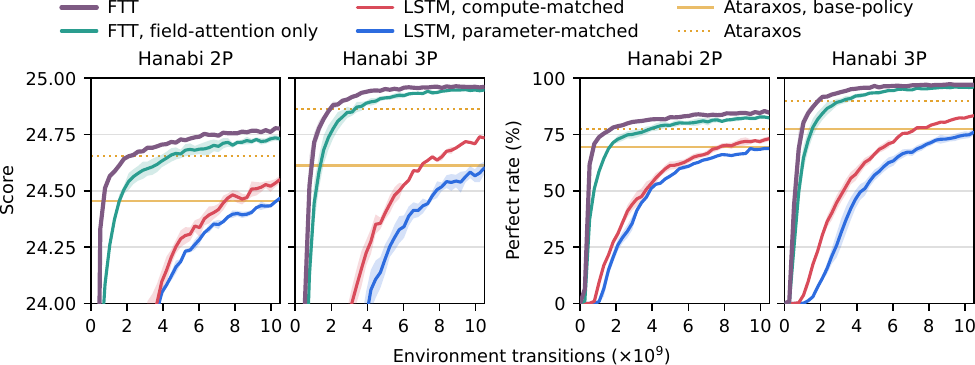}
\caption{Different architectures in Hanabi. The shaded area presents the uncertainty of $1.0 \times \mathrm{SEM}$.}
\label{fig:hanabi}
\end{figure}

We train \FTT\ agents on Hanabi for 2 to 5 players (Hanabi 2P to Hanabi 5P) using VRPO with a total of 40{,}000 iterations, with each iteration using 512 environments in parallel for 512 steps, forming a training schedule of approximately $1.05 \times 10^{10}$ timesteps. Each setting is trained with $3$ random seeds, and we report the averaged results in \Cref{tab:hanabi-published}. Compared to the previous results, the \FTT\ agent achieves the highest score and the highest perfect rate at every player count, even including those methods such as SPARTA~\citep{lerer2019sparta} and Ataraxos~\citep{sokota2026ataraxos} that are equipped with test-time improvement routines. Among all the settings, the \FTT\ agent has the greatest improvement in 5-player Hanabi, improving the results to 24.864 from 24.410 achieved by Ataraxos and playing a perfect game in about 89\% of the games, improved from 58\%.

\paragraph{Architectural Ablations.}

We study the performance of different architectures in 2- and 3-player Hanabi. Each model is trained with the same centralized critic but with different actor architectures. Each ablated model is trained under the same schedule of our main line \FTT, with either the parameters or the FLOPs per forward matching the main model. Field-attention-only \FTT\ removes the temporal attention from every block and moves the width to field attention so that both the parameters and the FLOPs per forward match the main model. Since the encoding writes the results of the hints into the cards, the field-attention-only \FTT\ keeps what the hints revealed but cannot see the order of the hints and the card plays. We also compare the model with \emph{LSTM}, which is the standard actor of the Hanabi literature~\citep{hu2019sad,hu2020otherplay,hu2021offbelief}: The model operates on an observation of one flat one-hot vector with two LSTM layers and a projection readout. The feature dimensions of the LSTM are chosen to approximately match the number of parameters or FLOPs used per timestep, resulting in feature dimensions of 1280 and 384, respectively.

\begin{table}[!ht]
\centering\small
\begin{tabular}{@{}lrr@{}}
\toprule
Actor Architecture & Parameters & FLOPs per forward \\
\midrule
\textbf{\FTT} & 2.56M & 58.8M \\
\FTT, field-attention only & 2.55M & 58.0M \\
LSTM, compute-matched (dim 1280) & 27.70M & 55.4M \\
LSTM, parameter-matched (dim 384) & 2.81M & 5.6M \\
\bottomrule
\end{tabular}

\caption{Parameters and FLOPs per forward of the actor architectures in 2-player Hanabi.}
\label{tab:hanabi-budgets}
\end{table}

\begin{figure}[t]
\centering
\includegraphics[width=\linewidth]{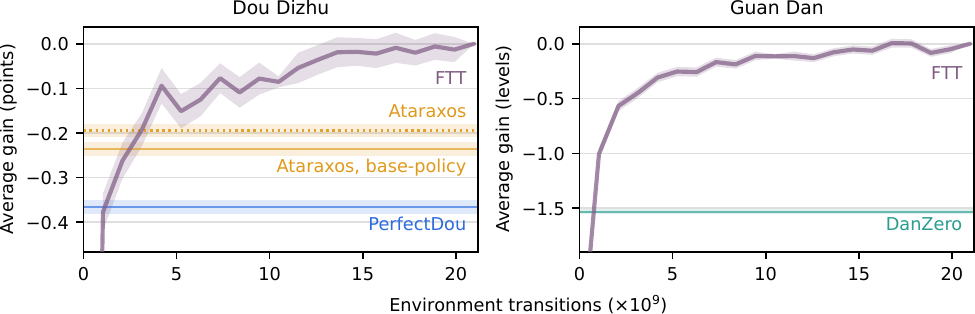}
\caption{Average gain against the final \FTT\ policy during training in Dou Dizhu and Guan Dan, where the final policy's comparison with itself is zero by definition. Horizontal lines present the performance of the released agents against the same final policy. The shaded area presents the uncertainty of $1.0 \times \mathrm{SEM}$.}
\label{fig:cards}
\end{figure}

We track the score and the perfect rate during training, and present the result in \Cref{fig:hanabi}. It can be seen that the \FTT\ agent has a much faster learning curve as well as a higher final score. After 40{,}000 iterations, the \FTT\ agent plays a perfect game in $84.7\%$ of the 2-player and $97.1\%$ of the 3-player games, against $73.2\%$ and $83.4\%$ for the compute-matched LSTM, which has more than ten times as many parameters. In 2-player Hanabi, the \FTT\ agent surpasses the base policy of Ataraxos after only 3{,}000 iterations, roughly $7.9 \times 10^{8}$ timesteps. The field-attention-only \FTT\ already learns faster and scores higher than both LSTM actors, and the temporal attention further improves both the learning speed and the final performance.

\subsection{Dou Dizhu, Guan Dan, and Dark Chess}
\label{sec:competitive}

We also apply \FTT\ to several competitive multi-player imperfect-information games and compare it with existing agents. Each \FTT\ agent is trained using VRPO for 40{,}000 iterations, with each iteration using 1{,}024 environments in parallel for 512 steps, forming a training schedule of approximately $2.10 \times 10^{10}$ timesteps.

\paragraph{Dou Dizhu.}
Dou Dizhu is a three-player card game between one landlord and two peasants using a 54-card poker deck. A landlord is decided by a bidding phase at game start and plays against the other two players. The goal of the players is to discard all their hands first within a set of available patterns in the subsequent playing phase.

We treat the card rank as fields in Dou Dizhu. This results in 15 local fields containing 13 normal poker ranks and two additional joker ranks. At each time step, each local field records how many cards are discarded by every player, held by the hero player, and revealed as the landlord's extra cards. The actions are split into multiple stages of decisions so that the kickers are decided in auxiliary timesteps. This allows us to restrict the main actions to solo, pair, three of a kind, straights, etc., each read out from the field of its main rank through its own output. In this way, \FTT\ encodes the large action set of more than 27{,}000 distinct card combinations without a flat output layer over all of them.

We evaluate the performance of the \FTT\ agent using the fixed-landlord card-play benchmark according to PerfectDou~\citep{yang2022perfectdou} with only the playing phase considered: each deal is played twice with the roles exchanged, where the landlord receives $\pm2$ points, doubled for every bomb. The performance is then evaluated by the average gain over the two role assignments.

The \FTT\ agent presents a strong performance under this benchmark. It surpasses the previous best agent, Ataraxos, in 6{,}000 iterations, or roughly $3.1 \times 10^{9}$ timesteps. Under an evaluation of 10{,}000 duplicate deals, the \FTT\ agent presents an average gain of $0.236\pm0.015$ against the base policy of Ataraxos~\citep{sokota2026ataraxos}, $0.366\pm0.016$ against PerfectDou~\citep{yang2022perfectdou}, and $0.548\pm0.016$ against DouZero~\citep{zha2021douzero}. It also beats Ataraxos by $0.194\pm0.015$ points with the test-time policy improvement of Ataraxos enabled, at a nominal rollout budget of 200 per decision and an update-equivalent step size of 5.

\paragraph{Guan Dan.}
Guan Dan is a four-player card game between two teams using two poker decks of a total of 108 cards. Similar to Dou Dizhu, the goal of the players is to discard all their hands before the opponent team. The game has a leveling on top of the normal gameplay structure in which winning a hand proceeds to the next level, and the team reaching the final level wins.

We treat the card rank as fields in Guan Dan similar to Dou Dizhu. This again results in 15 local fields containing 13 normal poker ranks and two additional joker ranks. Since the suit in Guan Dan has functionality for feasible patterns, each local field records how many cards are discarded by every player with different suits counted separately. As there are wild cards in Guan Dan, the actions are split into two stages, where in the first stage the pattern of the formed action is decided, and in the second stage the individual cards are selected to fulfill the pattern.

We evaluate the performance of the \FTT\ agent in single hands at the first level. The \FTT\ agent gains $1.532\pm0.023$ levels and a win rate of $72.7\%$ per hand against DanZero~\citep{lu2023danzero} over 2,000 duplicate deals.

\begin{figure}[t]
\centering
\begin{minipage}[c]{0.46\linewidth}
\includegraphics[width=\linewidth]{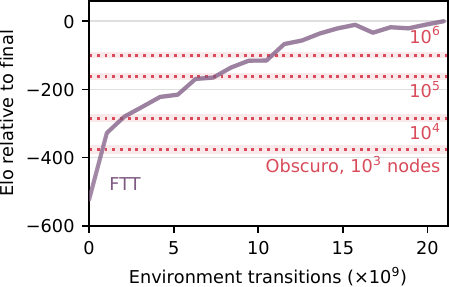}
\end{minipage}\hfill
\begin{minipage}[c]{0.5\linewidth}
\caption{Dark Chess. Elo of the \FTT\ checkpoints during training relative to the final checkpoint, from a round robin among them, and of our reimplementation of the search of Obscuro at four tree sizes (dotted), fitted into the same ratings from its matches against the checkpoints (Appendix~\ref{app:evaluation}). The shaded area presents the uncertainty of $1.0 \times \mathrm{SEM}$.}
\label{fig:darkchess-curves}
\end{minipage}
\end{figure}

\paragraph{Dark Chess.}
Dark Chess is a chess variant with fog of war: the players can only see their own pieces and the squares they can move to. This allows bluffing and requires the players to anticipate threats they cannot see.

We treat the 64 squares of the board as the local fields in Dark Chess. At each time step, each square field holds a single token for what the player sees there: whether it is a hidden square, an empty square, a self piece, a visible enemy piece, or an invisible square of the enemy that blocks the pawn. The global field further contains the additional information, including the player's color, whether it is to move, the castling rights and the move numbers. A move is read out from the field of its source square, with one output for each displacement, knight move and promotion.

We evaluate the \FTT\ agent with a round robin among the checkpoints of its run and compare it with the search of Obscuro~\citep{zhang2026obscuro} in \cref{fig:darkchess-curves}. As no code of Obscuro is published, we reimplemented its search with a budget in nodes of the search tree; at its published setting of 5 seconds per move, Obscuro searches 1,070,552 nodes on average. With one forward pass per decision, the final \FTT\ policy scores above one half against every tree size from 1,000 to 3,000,000 nodes, including $59.6\pm2.0\%$ against 1,000,000 nodes, a tree size that the training curve passes after about 21,000 iterations. Appendix~\ref{app:evaluation} describes the reimplementation, and Appendix~\ref{app:darkchess} lists all matches.

\section{Conclusion}
\label{sec:conclusion}

The Field-Time Transformer treats the observation history of an imperfect-information game as a grid of fields over time, attends across the fields of each observation and along the history of each field, and reads every action from the field it belongs to. With one forward pass per decision, the \FTT\ agent scores above the published Hanabi results at every player count, including the results of agents that use test-time policy improvement, and in five-player Hanabi it raises the perfect rate from $58\%$ to $89\%$. The ablations attribute this to the architecture. Sharing one set of weights across the fields of the observation makes learning faster and the final policy stronger than a recurrent actor on a flat vector, and an LSTM with the same compute per decision and more than ten times as many parameters does not close the gap. Letting every field attend over its own past then adds several points of perfect rate. The same organization transfers to the large games. With only the input tokens and the readouts specific to the game, the \FTT\ agents beat the released agents in Dou Dizhu and Guan Dan, among them Ataraxos with its test-time policy improvement enabled. In Dark Chess, the \FTT\ agent scores above one half against our reimplementation of the search of Obscuro at every tree size from 1,000 to 3,000,000 nodes, a range that contains the tree size of Obscuro's published setting.

\bibliographystyle{plainnat}
\bibliography{references}

\begin{thebibliography}{48}
\providecommand{\natexlab}[1]{#1}
\providecommand{\url}[1]{\texttt{#1}}
\expandafter\ifx\csname urlstyle\endcsname\relax
  \providecommand{\doi}[1]{doi: #1}\else
  \providecommand{\doi}{doi: \begingroup \urlstyle{rm}\Url}\fi

\bibitem[Arnab et~al.(2021)Arnab, Dehghani, Heigold, Sun, Lucic, and Schmid]{arnab2021vivit}
Anurag Arnab, Mostafa Dehghani, Georg Heigold, Chen Sun, Mario Lucic, and Cordelia Schmid.
\newblock {ViViT}: A video vision transformer.
\newblock In \emph{Proceedings of the IEEE/CVF International Conference on Computer Vision}, 2021.
\newblock URL \url{https://openaccess.thecvf.com/content/ICCV2021/html/Arnab_ViViT_A_Video_Vision_Transformer_ICCV_2021_paper.html}.

\bibitem[Bard et~al.(2019)Bard, Foerster, Chandar, Burch, Lanctot, Song, Parisotto, Dumoulin, Moitra, Hughes, Dunning, Mourad, Larochelle, Bellemare, and Bowling]{bard2019hanabi}
Nolan Bard, Jakob~N. Foerster, Sarath Chandar, Neil Burch, Marc Lanctot, H.~Francis Song, Emilio Parisotto, Vincent Dumoulin, Subhodeep Moitra, Edward Hughes, Iain Dunning, Shibl Mourad, Hugo Larochelle, Marc~G. Bellemare, and Michael Bowling.
\newblock The hanabi challenge: A new frontier for {AI} research, 2019.
\newblock URL \url{https://arxiv.org/abs/1902.00506}.

\bibitem[Beltagy et~al.(2020)Beltagy, Peters, and Cohan]{beltagy2020longformer}
Iz~Beltagy, Matthew~E. Peters, and Arman Cohan.
\newblock Longformer: The long-document transformer, 2020.

\bibitem[Berner et~al.(2019)Berner, Brockman, Chan, Cheung, D{\k{e}}biak, Dennison, Farhi, Fischer, Hashme, Hesse, et~al.]{berner2019dota}
Christopher Berner, Greg Brockman, Brooke Chan, Vicki Cheung, Przemys{\l}aw D{\k{e}}biak, Christy Dennison, David Farhi, Quirin Fischer, Shariq Hashme, Chris Hesse, et~al.
\newblock Dota 2 with large scale deep reinforcement learning, 2019.
\newblock URL \url{https://arxiv.org/abs/1912.06680}.

\bibitem[Bertasius et~al.(2021)Bertasius, Wang, and Torresani]{bertasius2021timesformer}
Gedas Bertasius, Heng Wang, and Lorenzo Torresani.
\newblock Is space-time attention all you need for video understanding?
\newblock In \emph{Proceedings of the 38th International Conference on Machine Learning}, volume 139 of \emph{Proceedings of Machine Learning Research}, pages 813--824, 2021.
\newblock URL \url{https://proceedings.mlr.press/v139/bertasius21a.html}.

\bibitem[Blattmann et~al.(2023)Blattmann, Rombach, Ling, Dockhorn, Kim, Fidler, and Kreis]{blattmann2023align}
Andreas Blattmann, Robin Rombach, Huan Ling, Tim Dockhorn, Seung~Wook Kim, Sanja Fidler, and Karsten Kreis.
\newblock Align your latents: High-resolution video synthesis with latent diffusion models.
\newblock In \emph{Proceedings of the IEEE/CVF Conference on Computer Vision and Pattern Recognition}, 2023.
\newblock \doi{10.1109/CVPR52729.2023.02161}.
\newblock URL \url{https://doi.org/10.1109/CVPR52729.2023.02161}.

\bibitem[Bruce et~al.(2024)Bruce, Dennis, Edwards, Parker-Holder, Shi, Hughes, Lai, Mavalankar, Steigerwald, Apps, et~al.]{bruce2024genie}
Jake Bruce, Michael~D. Dennis, Ashley Edwards, Jack Parker-Holder, Yuge Shi, Edward Hughes, Matthew Lai, Aditi Mavalankar, Richie Steigerwald, Chris Apps, et~al.
\newblock {Genie}: Generative interactive environments.
\newblock In \emph{Proceedings of the 41st International Conference on Machine Learning}, volume 235 of \emph{Proceedings of Machine Learning Research}, pages 4603--4623, 2024.
\newblock URL \url{https://proceedings.mlr.press/v235/bruce24a.html}.

\bibitem[Fan et~al.(2021)Fan, Xiong, Mangalam, Li, Yan, Malik, and Feichtenhofer]{fan2021mvit}
Haoqi Fan, Bo~Xiong, Karttikeya Mangalam, Yanghao Li, Zhicheng Yan, Jitendra Malik, and Christoph Feichtenhofer.
\newblock Multiscale vision transformers.
\newblock In \emph{Proceedings of the IEEE/CVF International Conference on Computer Vision}, 2021.
\newblock \doi{10.1109/ICCV48922.2021.00675}.
\newblock URL \url{https://doi.org/10.1109/ICCV48922.2021.00675}.

\bibitem[Fan and Farina(2026)]{fan2026vrpo}
Zhiyuan Fan and Gabriele Farina.
\newblock {GAE} falls short in imperfect-information self-play reinforcement learning, 2026.
\newblock URL \url{https://arxiv.org/abs/2605.19235}.

\bibitem[Forkel et~al.(2025)Forkel, Ruhdorfer, Beukman, Bulling, and Foerster]{forkel2025entropy}
Johannes Forkel, Constantin Ruhdorfer, Michael Beukman, Andreas Bulling, and Jakob Foerster.
\newblock High entropy leads to symmetry-equivariant policies in {Dec-POMDPs}, 2025.
\newblock URL \url{https://arxiv.org/abs/2511.22581}.

\bibitem[Fu et~al.(2022)Fu, Liu, Wu, Wang, Yang, Li, Xing, Li, Ma, Fu, and Wei]{fu2022ach}
Haobo Fu, Weiming Liu, Shuang Wu, Yijia Wang, Tao Yang, Kai Li, Junliang Xing, Bin Li, Bo~Ma, Qiang Fu, and Yang Wei.
\newblock Actor-critic policy optimization in a large-scale imperfect-information game.
\newblock In \emph{International Conference on Learning Representations}, 2022.
\newblock URL \url{https://openreview.net/forum?id=DTXZqTNV5nW}.

\bibitem[Girdhar and Grauman(2021)]{girdhar2021avt}
Rohit Girdhar and Kristen Grauman.
\newblock Anticipative video transformer.
\newblock In \emph{Proceedings of the IEEE/CVF International Conference on Computer Vision}, 2021.
\newblock \doi{10.1109/ICCV48922.2021.01325}.
\newblock URL \url{https://doi.org/10.1109/ICCV48922.2021.01325}.

\bibitem[Ho et~al.(2022)Ho, Salimans, Gritsenko, Chan, Norouzi, and Fleet]{ho2022vdm}
Jonathan Ho, Tim Salimans, Alexey Gritsenko, William Chan, Mohammad Norouzi, and David~J. Fleet.
\newblock Video diffusion models.
\newblock In \emph{Advances in Neural Information Processing Systems}, 2022.
\newblock \doi{10.52202/068431-0628}.
\newblock URL \url{https://doi.org/10.52202/068431-0628}.

\bibitem[Hochreiter and Schmidhuber(1997)]{hochreiter1997lstm}
Sepp Hochreiter and J{\"u}rgen Schmidhuber.
\newblock Long short-term memory.
\newblock \emph{Neural Computation}, 9\penalty0 (8):\penalty0 1735--1780, 1997.

\bibitem[Hu and Foerster(2019)]{hu2019sad}
Hengyuan Hu and Jakob~N. Foerster.
\newblock Simplified action decoder for deep multi-agent reinforcement learning, 2019.
\newblock URL \url{https://arxiv.org/abs/1912.02288}.

\bibitem[Hu et~al.(2020)Hu, Lerer, Peysakhovich, and Foerster]{hu2020otherplay}
Hengyuan Hu, Adam Lerer, Alex Peysakhovich, and Jakob Foerster.
\newblock ``other-play'' for zero-shot coordination.
\newblock In \emph{Proceedings of the 37th International Conference on Machine Learning}, volume 119 of \emph{Proceedings of Machine Learning Research}, pages 4399--4410, 2020.
\newblock URL \url{https://proceedings.mlr.press/v119/hu20a.html}.

\bibitem[Hu et~al.(2021)Hu, Lerer, Cui, Pineda, Brown, and Foerster]{hu2021offbelief}
Hengyuan Hu, Adam Lerer, Brandon Cui, Luis Pineda, Noam Brown, and Jakob Foerster.
\newblock Off-belief learning.
\newblock In \emph{Proceedings of the 38th International Conference on Machine Learning}, volume 139 of \emph{Proceedings of Machine Learning Research}, pages 4369--4379, 2021.
\newblock URL \url{https://proceedings.mlr.press/v139/hu21c.html}.

\bibitem[Jordan et~al.(2024)Jordan, Jin, Boza, You, Cesista, Newhouse, and Bernstein]{jordan2024muon}
Keller Jordan, Yuchen Jin, Vlado Boza, Jiacheng You, Franz Cesista, Laker Newhouse, and Jeremy Bernstein.
\newblock Muon: An optimizer for hidden layers in neural networks, 2024.
\newblock URL \url{https://kellerjordan.github.io/posts/muon/}.

\bibitem[Lerer et~al.(2019)Lerer, Hu, Foerster, and Brown]{lerer2019sparta}
Adam Lerer, Hengyuan Hu, Jakob Foerster, and Noam Brown.
\newblock Improving policies via search in cooperative partially observable games, 2019.
\newblock URL \url{https://arxiv.org/abs/1912.02318}.

\bibitem[Li et~al.(2020)Li, Koyamada, Ye, Liu, Wang, Yang, Zhao, Qin, Liu, and Hon]{li2020suphx}
Junjie Li, Sotetsu Koyamada, Qiwei Ye, Guoqing Liu, Chao Wang, Ruihan Yang, Li~Zhao, Tao Qin, Tie-Yan Liu, and Hsiao-Wuen Hon.
\newblock {Suphx}: Mastering mahjong with deep reinforcement learning, 2020.
\newblock URL \url{https://arxiv.org/abs/2003.13590}.

\bibitem[Liu et~al.(2022)Liu, Ning, Cao, Wei, Zhang, Lin, and Hu]{liu2022videoswin}
Ze~Liu, Jia Ning, Yue Cao, Yixuan Wei, Zheng Zhang, Stephen Lin, and Han Hu.
\newblock Video {Swin} transformer.
\newblock In \emph{Proceedings of the IEEE/CVF Conference on Computer Vision and Pattern Recognition}, 2022.
\newblock \doi{10.1109/CVPR52688.2022.00320}.
\newblock URL \url{https://doi.org/10.1109/CVPR52688.2022.00320}.

\bibitem[Lu et~al.(2022)Lu, Zhao, Zhao, Zhou, and Li]{lu2023danzero}
Yudong Lu, Jian Zhao, Youpeng Zhao, Wengang Zhou, and Houqiang Li.
\newblock {DanZero}: Mastering {GuanDan} game with reinforcement learning, 2022.
\newblock URL \url{https://arxiv.org/abs/2210.17087}.

\bibitem[Ma et~al.(2025)Ma, Wang, Chen, Jia, Liu, Li, Chen, and Qiao]{ma2025latte}
Xin Ma, Yaohui Wang, Xinyuan Chen, Gengyun Jia, Ziwei Liu, Yuan-Fang Li, Cunjian Chen, and Yu~Qiao.
\newblock {Latte}: Latent diffusion transformer for video generation.
\newblock \emph{Transactions on Machine Learning Research}, 2025.
\newblock ISSN 2835-8856.
\newblock URL \url{https://openreview.net/forum?id=ntGPYNUF3t}.

\bibitem[Neimark et~al.(2021)Neimark, Bar, Zohar, and Asselmann]{neimark2021vtn}
Daniel Neimark, Omri Bar, Maya Zohar, and Dotan Asselmann.
\newblock Video transformer network.
\newblock In \emph{Proceedings of the IEEE/CVF International Conference on Computer Vision Workshops}, 2021.
\newblock \doi{10.1109/ICCVW54120.2021.00355}.
\newblock URL \url{https://doi.org/10.1109/ICCVW54120.2021.00355}.

\bibitem[Ngiam et~al.(2022)Ngiam, Caine, Vasudevan, Zhang, Chiang, Ling, Roelofs, Bewley, Liu, Venugopal, Weiss, Sapp, Chen, and Shlens]{ngiam2022scene}
Jiquan Ngiam, Benjamin Caine, Vijay Vasudevan, Zhengdong Zhang, Hao-Tien~Lewis Chiang, Jeffrey Ling, Rebecca Roelofs, Alex Bewley, Chenxi Liu, Ashish Venugopal, David Weiss, Ben Sapp, Zhifeng Chen, and Jonathon Shlens.
\newblock Scene transformer: A unified architecture for predicting multiple agent trajectories.
\newblock In \emph{International Conference on Learning Representations}, 2022.
\newblock URL \url{https://arxiv.org/abs/2106.08417}.

\bibitem[Patrick et~al.(2021)Patrick, Campbell, Asano, Misra, Metze, Feichtenhofer, Vedaldi, and Henriques]{patrick2021motionformer}
Mandela Patrick, Dylan Campbell, Yuki Asano, Ishan Misra, Florian Metze, Christoph Feichtenhofer, Andrea Vedaldi, and Jo{\~a}o~F. Henriques.
\newblock Keeping your eye on the ball: Trajectory attention in video transformers.
\newblock In \emph{Advances in Neural Information Processing Systems}, 2021.
\newblock URL \url{https://arxiv.org/abs/2106.05392}.

\bibitem[Perolat et~al.(2022)Perolat, De~Vylder, Hennes, Tarassov, Strub, de~Boer, Muller, Connor, Burch, Anthony, et~al.]{perolat2022mastering}
Julien Perolat, Bart De~Vylder, Daniel Hennes, Eugene Tarassov, Florian Strub, Vincent de~Boer, Paul Muller, Jerome~T. Connor, Neil Burch, Thomas Anthony, et~al.
\newblock Mastering the game of {Stratego} with model-free multiagent reinforcement learning.
\newblock \emph{Science}, 378\penalty0 (6623):\penalty0 990--996, 2022.

\bibitem[Qiu et~al.(2017)Qiu, Yao, and Mei]{qiu2017pseudo3d}
Zhaofan Qiu, Ting Yao, and Tao Mei.
\newblock Learning spatio-temporal representation with pseudo-{3D} residual networks.
\newblock In \emph{Proceedings of the IEEE International Conference on Computer Vision}, 2017.
\newblock \doi{10.1109/ICCV.2017.590}.
\newblock URL \url{https://doi.org/10.1109/ICCV.2017.590}.

\bibitem[Schulman et~al.(2015)Schulman, Moritz, Levine, Jordan, and Abbeel]{schulman2015gae}
John Schulman, Philipp Moritz, Sergey Levine, Michael Jordan, and Pieter Abbeel.
\newblock High-dimensional continuous control using generalized advantage estimation, 2015.
\newblock URL \url{https://arxiv.org/abs/1506.02438}.

\bibitem[Schulman et~al.(2017)Schulman, Wolski, Dhariwal, Radford, and Klimov]{schulman2017ppo}
John Schulman, Filip Wolski, Prafulla Dhariwal, Alec Radford, and Oleg Klimov.
\newblock Proximal policy optimization algorithms, 2017.
\newblock URL \url{https://arxiv.org/abs/1707.06347}.

\bibitem[Shazeer(2020)]{shazeer2020glu}
Noam Shazeer.
\newblock {GLU} variants improve transformer, 2020.
\newblock URL \url{https://arxiv.org/abs/2002.05202}.

\bibitem[Singer et~al.(2023)Singer, Polyak, Hayes, Yin, An, Zhang, Hu, Yang, Ashual, Gafni, Parikh, Gupta, and Taigman]{singer2023makeavideo}
Uriel Singer, Adam Polyak, Thomas Hayes, Xi~Yin, Jie An, Songyang Zhang, Qiyuan Hu, Harry Yang, Oron Ashual, Oran Gafni, Devi Parikh, Sonal Gupta, and Yaniv Taigman.
\newblock {Make-A-Video}: Text-to-video generation without text-video data.
\newblock In \emph{International Conference on Learning Representations}, 2023.
\newblock URL \url{https://openreview.net/forum?id=nJfylDvgzlq}.

\bibitem[Sokota et~al.(2022)Sokota, D'Orazio, Kolter, Loizou, Lanctot, Mitliagkas, Brown, and Kroer]{sokota2022unified}
Samuel Sokota, Ryan D'Orazio, J.~Zico Kolter, Nicolas Loizou, Marc Lanctot, Ioannis Mitliagkas, Noam Brown, and Christian Kroer.
\newblock A unified approach to reinforcement learning, quantal response equilibria, and two-player zero-sum games, 2022.
\newblock URL \url{https://arxiv.org/abs/2206.05825}.

\bibitem[Sokota et~al.(2026)Sokota, Vinitsky, Hu, Fan, Kolter, and Farina]{sokota2026ataraxos}
Samuel Sokota, Eugene Vinitsky, Hengyuan Hu, Zhiyuan Fan, J.~Zico Kolter, and Gabriele Farina.
\newblock Scalable decision-making for games of imperfect information.
\newblock \emph{Nature}, 658:\penalty0 55--59, 2026.
\newblock \doi{10.1038/s41586-026-11036-y}.
\newblock URL \url{https://www.nature.com/articles/s41586-026-11036-y}.

\bibitem[Straka et~al.(2026)Straka, Lis{\'y}, and Schmid]{straka2026generals}
Matej Straka, Viliam Lis{\'y}, and Martin Schmid.
\newblock Superhuman {AI} for {Generals.io} using self-play reinforcement learning, 2026.
\newblock URL \url{https://arxiv.org/abs/2606.23348}.

\bibitem[Su et~al.(2021)Su, Lu, Pan, Murtadha, Wen, and Liu]{su2021roformer}
Jianlin Su, Yu~Lu, Shengfeng Pan, Ahmed Murtadha, Bo~Wen, and Yunfeng Liu.
\newblock {RoFormer}: Enhanced transformer with rotary position embedding, 2021.
\newblock URL \url{https://arxiv.org/abs/2104.09864}.

\bibitem[{TechNode}(2023)]{tencent2023luckyj}
{TechNode}.
\newblock Tencent's mahjong {AI} sets new gaming record on international mahjong platform.
\newblock TechNode, 12 July 2023, 2023.
\newblock URL \url{https://technode.com/2023/07/12/tencents-mahjong-ai-sets-new-gaming-record-on-international-mahjong-platform}.

\bibitem[Tran et~al.(2018)Tran, Wang, Torresani, Ray, LeCun, and Paluri]{tran2018closer}
Du~Tran, Heng Wang, Lorenzo Torresani, Jamie Ray, Yann LeCun, and Manohar Paluri.
\newblock A closer look at spatiotemporal convolutions for action recognition.
\newblock In \emph{Proceedings of the IEEE/CVF Conference on Computer Vision and Pattern Recognition}, 2018.
\newblock \doi{10.1109/CVPR.2018.00675}.
\newblock URL \url{https://doi.org/10.1109/CVPR.2018.00675}.

\bibitem[Vaswani et~al.(2017)Vaswani, Shazeer, Parmar, Uszkoreit, Jones, Gomez, Kaiser, and Polosukhin]{vaswani2017attention}
Ashish Vaswani, Noam Shazeer, Niki Parmar, Jakob Uszkoreit, Llion Jones, Aidan~N. Gomez, Lukasz Kaiser, and Illia Polosukhin.
\newblock Attention is all you need.
\newblock In \emph{Advances in Neural Information Processing Systems}, 2017.
\newblock URL \url{https://arxiv.org/abs/1706.03762}.

\bibitem[Villegas et~al.(2023)Villegas, Babaeizadeh, Kindermans, Moraldo, Zhang, Saffar, Castro, Kunze, and Erhan]{villegas2023phenaki}
Ruben Villegas, Mohammad Babaeizadeh, Pieter-Jan Kindermans, Hernan Moraldo, Han Zhang, Mohammad~Taghi Saffar, Santiago Castro, Julius Kunze, and Dumitru Erhan.
\newblock {Phenaki}: Variable length video generation from open domain textual descriptions.
\newblock In \emph{International Conference on Learning Representations}, 2023.
\newblock URL \url{https://openreview.net/forum?id=vOEXS39nOF}.

\bibitem[Vinyals et~al.(2019)Vinyals, Babuschkin, Czarnecki, Mathieu, Dudzik, Chung, Choi, Powell, Ewalds, Georgiev, et~al.]{vinyals2019grandmaster}
Oriol Vinyals, Igor Babuschkin, Wojciech~M. Czarnecki, Micha{\"e}l Mathieu, Andrew Dudzik, Junyoung Chung, David~H. Choi, Richard Powell, Timo Ewalds, Petko Georgiev, et~al.
\newblock Grandmaster level in {StarCraft II} using multi-agent reinforcement learning.
\newblock \emph{Nature}, 575\penalty0 (7782):\penalty0 350--354, 2019.

\bibitem[Xie et~al.(2018)Xie, Sun, Huang, Tu, and Murphy]{xie2018rethinking}
Saining Xie, Chen Sun, Jonathan Huang, Zhuowen Tu, and Kevin Murphy.
\newblock Rethinking spatiotemporal feature learning: Speed-accuracy trade-offs in video classification.
\newblock In \emph{Proceedings of the European Conference on Computer Vision}, 2018.
\newblock \doi{10.1007/978-3-030-01267-0_19}.
\newblock URL \url{https://doi.org/10.1007/978-3-030-01267-0_19}.

\bibitem[Yang et~al.(2022)Yang, Liu, Hong, Zhang, Fang, Zeng, and Lin]{yang2022perfectdou}
Guan Yang, Minghuan Liu, Weijun Hong, Weinan Zhang, Fei Fang, Guangjun Zeng, and Yue Lin.
\newblock {PerfectDou}: Dominating {DouDizhu} with perfect information distillation.
\newblock In \emph{Advances in Neural Information Processing Systems}, 2022.
\newblock URL \url{https://arxiv.org/abs/2203.16406}.

\bibitem[Yu et~al.(2022)Yu, Velu, Vinitsky, Gao, Wang, Bayen, and Wu]{yu2022mappo}
Chao Yu, Akash Velu, Eugene Vinitsky, Jiaxuan Gao, Yu~Wang, Alexandre Bayen, and Yi~Wu.
\newblock The surprising effectiveness of {PPO} in cooperative, multi-agent games.
\newblock In \emph{Advances in Neural Information Processing Systems, Datasets and Benchmarks Track}, 2022.
\newblock URL \url{https://arxiv.org/abs/2103.01955}.

\bibitem[Zha et~al.(2021)Zha, Xie, Ma, Zhang, Lian, Hu, and Liu]{zha2021douzero}
Daochen Zha, Jingru Xie, Wenye Ma, Sheng Zhang, Xiangru Lian, Xia Hu, and Ji~Liu.
\newblock {DouZero}: Mastering {DouDizhu} with self-play deep reinforcement learning.
\newblock In \emph{Proceedings of the 38th International Conference on Machine Learning}, volume 139 of \emph{Proceedings of Machine Learning Research}, pages 12333--12344, 2021.
\newblock URL \url{https://proceedings.mlr.press/v139/zha21a.html}.

\bibitem[Zhang and Sennrich(2019)]{zhang2019rmsnorm}
Biao Zhang and Rico Sennrich.
\newblock Root mean square layer normalization.
\newblock In \emph{Advances in Neural Information Processing Systems}, 2019.
\newblock URL \url{https://arxiv.org/abs/1910.07467}.

\bibitem[Zhang and Sandholm(2026)]{zhang2026obscuro}
Brian~Hu Zhang and Tuomas Sandholm.
\newblock General search techniques without common knowledge for imperfect-information games, and application to superhuman fog of war chess.
\newblock In \emph{International Conference on Learning Representations}, 2026.
\newblock URL \url{https://arxiv.org/abs/2506.01242}.

\bibitem[Zhang et~al.(2021)Zhang, Li, Liu, Shuai, Zhu, Brattoli, Chen, Marsic, and Tighe]{zhang2021vidtr}
Yanyi Zhang, Xinyu Li, Chunhui Liu, Bing Shuai, Yi~Zhu, Biagio Brattoli, Hao Chen, Ivan Marsic, and Joseph Tighe.
\newblock {VidTr}: Video transformer without convolutions.
\newblock In \emph{Proceedings of the IEEE/CVF International Conference on Computer Vision}, 2021.
\newblock \doi{10.1109/ICCV48922.2021.01332}.
\newblock URL \url{https://doi.org/10.1109/ICCV48922.2021.01332}.

\end{thebibliography}
\clearpage
\appendix
{\LARGE Appendix}

\section{Games Description}
\label{app:games}

\paragraph{Hanabi.}
Hanabi is a cooperative card game for two to five players~\citep{bard2019hanabi}. The deck has 50 cards in five colors, and each color has three cards of rank 1, two each of ranks 2 to 4 and one of rank 5. A player holds five cards in games of two or three players and four cards in games of four or five, and sees every hand except its own. On its turn a player gives a hint, discards a card or plays a card. A hint names a color or a rank, points out all cards of that color or rank in one teammate's hand and costs one of the eight hint tokens. A discard returns a hint token. A played card has to be the next rank of its color's firework; otherwise the team loses one of its three lives. Completing a firework with its rank-5 card also returns a hint token. The game ends when all five fireworks are complete, when the lives are used up, or one round after the deck runs out. The score is the number of cards played, at most 25, and it is zero when all lives are lost. We use the games with two to five players (Hanabi 2P to Hanabi 5P).

\paragraph{Dou Dizhu.}
Dou Dizhu is a three-player shedding game with a 54-card deck, played by one landlord against two peasants. After the deal the players bid for the role of the landlord, who receives three extra cards that are shown to everyone, and the players may then double the stakes. In the playing phase the players take turns to play a combination of cards that beats the previous one, such as a single card, a pair, a triple with attached cards, a straight or a bomb, or to pass. The side of the first player to empty its hand wins. The stakes double with every bomb played. Dou Dizhu is a standard benchmark for self-play reinforcement learning~\citep{zha2021douzero,yang2022perfectdou}. Our comparisons use the card-play benchmark of PerfectDou, in which the landlord is fixed and only the playing phase is played.

\paragraph{Guan Dan.}
Guan Dan is a four-player shedding game between two teams of partners who sit opposite each other, played with two 54-card decks, 27 cards per player~\citep{lu2023danzero}. A hand is played at a level, which is one of the thirteen ranks. Cards of the level rank are higher than every other card except the jokers, and the two hearts of the level rank are wild cards. The combinations are singles, pairs, triples, full houses, five-card straights, three consecutive pairs, two consecutive triples, bombs of four to ten cards of one rank, straight flushes and the four jokers. The team of the first player to finish wins the hand and advances three levels if the partner finishes second, two if third and one if last. The next hand starts with a tribute, in which the losers give their highest cards to the winners, and a match is won by the team that reaches the final level. We train and evaluate on single hands.

\paragraph{Dark Chess.}
Dark Chess, also called fog-of-war chess, is chess in which a player sees only its own pieces and the squares to which they can move~\citep{zhang2026obscuro}. A pawn sees the square in front of it only when that square is empty and its capture squares only when they hold an enemy piece. When a pawn is blocked by a piece it cannot see, the player learns that the square is occupied and nothing more. There is no check: a move may leave the player's own king attacked, a player whose king can be captured loses, and so does a player without a move. Castling and en passant are available. A game is drawn by threefold repetition, by the fifty-move rule or after 1,000 plies.

\clearpage
\section{Game Encoding}
\label{app:interfaces}

Every entry of a field is one categorical token in \cref{eq:embedding}, and the local and global readouts are those of \cref{eq:head}. In \cref{tab:games}, \textbf{Field Count} is the number $K$ of local fields, \textbf{Token Slots} the number of entries of a local field and of the global field, \textbf{Total Actions} the size of the action set including \textsc{Noop}, and \textbf{Action Readouts} the outputs of a local field and of the global field.

\subsection{Hanabi}

\paragraph{Local Fields.}
The ten local fields are the five colors and the five ranks, and the two kinds share one slot format written along the other axis, the ranks of a color field and the colors of a rank field. A field records its progress, the height of that color's firework or the number of colors on which that rank has been played; which values of the other axis have been played and how many copies of each have been discarded; for every card in a partner's hand, the card's other-axis value if the card belongs to this field; and, for every card in every hand, whether the hints given so far allow the card to be of this color or rank. These possibility entries are those of the standard Hanabi observation and never reveal the identity of a card in the player's own hand.

\paragraph{Global Field.}
The global field holds the player's seat, whose turn it is, the hint and life tokens, the size of the deck, the end-game countdown, every hand's size and the last public action.

\paragraph{Action Readout.}
A hint is read from the field of the color or rank it names, with one output per partner, $P-1$ per local field. Playing or discarding a position and \textsc{Noop} are read from the global field, $2H+1$ outputs for a hand of $H$ cards.

\paragraph{Rewards.}
Every player receives the final score at the end of the game.

\subsection{Dou Dizhu}

\paragraph{Local Fields.}
The fifteen local fields are the thirteen ranks and the two jokers. A rank's field has six entries, each a count from zero to four: the cards of that rank in the player's hand, the cards of that rank played so far by each of the three players, those among the landlord's three revealed cards, and those already selected for the combination that the acting player is assembling.

\paragraph{Global Field.}
The global field has 30 entries, among them the phase of the game, the seats of the landlord and of the acting player, the bids and doublings, the combination to beat, the number of bombs played, the cards remaining in each hand, and the state of the combination being assembled.

\paragraph{Action Readout.}
A combination is built over consecutive decisions. Actions that name a rank use ten outputs of that rank's readout: a single, a pair, a triple or four of a kind of the rank; the start of a straight, of a chain of pairs or of a chain of triples at the rank; and the end of a chain at the rank, plain or with single or paired attachments. Passing, the rocket, the kind of attachment, bids, doubling and \textsc{Noop} use eleven outputs of the global readout. Not every output is meaningful for every rank, which leaves 134 actions in total.

\paragraph{Rewards.}
A peasant's stake is the bid, doubled for every bomb or rocket played, for a spring, in which the peasants have played no card or the landlord has played only its first combination, and for a doubling. Each peasant wins or loses its stake, and the landlord wins or loses the sum of the two.

\subsection{Guan Dan}

\paragraph{Local Fields.}
The fifteen local fields are again the ranks. Suits matter in Guan Dan, for straight flushes and for the wild cards of the current level, so a rank's field has 24 entries: for each of the four suits, the number of cards (zero to two, since two decks are used) in the player's hand, in the cards played by each of the four players, and in the combination being assembled.

\paragraph{Global Field.}
The global field has 34 entries, among them the level, the seats of the acting and the finished players, the combination to beat, the cards left in each hand and the tribute cards.

\paragraph{Action Readout.}
A combination is chosen in two stages, first its pattern and then the cards that fill it. The readout has thirteen outputs per rank and four for the global field: 172 actions.

\paragraph{Rewards.}
Each player of the winning team receives the number of levels won, three, two or one, and each player of the losing team the negative of it.

\subsection{Dark Chess}

\paragraph{Local Fields.}
The 64 local fields are the squares of the board, and the embedding of a field is the sum of a rank embedding and a file embedding. A square's field has a single entry with fifteen values: hidden, visibly empty, one of the twelve pieces, or an enemy piece of unknown type, which occurs when a pawn is blocked by a piece it cannot see.

\paragraph{Global Field.}
The global field has six entries: the player's color, whether it is to move, its two castling rights and the move number in two digits.

\paragraph{Action Readout.}
A move is read from the field of its source square, which has 76 outputs: 56 for a displacement of one to seven squares in one of eight directions, eight for the knight moves, and twelve for a promotion in one of three directions to one of four pieces. The 64 squares thus give 4,864 moves, and \textsc{Noop} is read from the global field.

\paragraph{Rewards.}
The winner receives $+1$ and the loser $-1$, and both players receive $0$ in a draw.

\clearpage
\section{Training Details}
\label{app:training}

\paragraph{Training Objective.}
All networks are trained by self-play with Variance-Reduced Policy Optimization (VRPO)~\citep{fan2026vrpo}, a PPO-style algorithm~\citep{schulman2017ppo} for stochastic self-play. VRPO trains a centralized action-value critic $Q(s,a)$ on the full state $s$ of the game, including the information hidden from the acting player, and estimates the advantage of the action taken at step $t$ by
\begin{align}
 \hat A_t&=Q(s_t,a_t)-\bar V(s_t)+\sum_{\tau\geq t}\lambda^{\tau-t}\,\delta_\tau,
 \label{eq:qboost}\\
 \bar V(s)&=\sum_{a}\pi(a\mid s)\,Q(s,a),
 \qquad
 \delta_\tau=r_\tau+\bar V(s_{\tau+1})-Q(s_\tau,a_\tau),\nonumber
\end{align}
where $\lambda$ is a trace parameter and $\pi(a\mid s)$ is the probability that the player who acts in $s$ chooses $a$. Taking the expectation over the policy in $\bar V$ removes the noise that sampled future actions add to generalized advantage estimation~\citep{schulman2015gae}, which matters when policies are as stochastic as they are in these games. Let $\pi^{\mathrm{old}}$ be the policy that collected the data of an iteration, $\rho_t=\pi(a_t\mid \vo_{1:t})/\pi^{\mathrm{old}}(a_t\mid \vo_{1:t})$ the probability ratio of the action taken, and $\tilde A_t$ the advantage estimate of \cref{eq:qboost} normalized over the minibatch to zero mean and unit variance. The policy-gradient part of the actor's objective is
\begin{equation}
 -\,\Expt_t\Bigl[\min\bigl\{\rho_t\,c_t,\;\mathrm{clip}(\rho_t,1-\eps,1+\eps)\,c_t\bigr\}\Bigr],
 \qquad c_t=\tilde A_t-\kappa\log\rho_t ,
\end{equation}
where $c_t$ is treated as a constant in the gradient; the term $-\kappa\log\rho_t$ in $c_t$ penalizes $\mathrm{KL}(\pi\,\|\,\pi^{\mathrm{old}})$ with coefficient $\kappa$. The critic is trained by regression of $Q(s_t,a_t)$ on the target $Q(s_t,a_t)+\sum_{\tau\geq t}\lambda^{\tau-t}\delta_\tau$ with the residuals $\delta_\tau$ of \cref{eq:qboost}, the target being held fixed. Rewards are the scores of the game at its end and are not discounted. \citet{fan2026vrpo} derive the estimator and analyze its variance.

\paragraph{Sequence Training.}
An iteration collects a fixed number of steps from parallel environments, so the collected fragments generally begin in the middle of an episode. The actor is trained on a fragment starting from the memory it had at the fragment's first step, and no gradient flows into that memory. Attention never crosses the boundary between two episodes inside a fragment.

\paragraph{Training Hyperparameters.}
We use Muon~\citep{jordan2024muon} with momentum $0.95$ and weight decay $0.01$ for the weight matrices of the attention projections, the feed-forward networks, the actor's readouts and the LSTM layers, and Adam with $\beta=(0.9,0.999)$ and no weight decay for the remaining parameters, namely the embedding tables, the gains of the normalizations, the biases and the output layer of the critic. Gradients are clipped to norm $1$. Every iteration performs one actor epoch and one critic epoch, each over 16 minibatches. The base clipping coefficient is $\eps_{\mathrm{base}}=0.2$, the reverse-KL coefficient is $\kappa=0.02$, and the trace parameter is $\lambda=0.95$. \Cref{tab:hyperparameters} lists the base learning rate $\eta$, the schedule constant $T$ and the training budget of each game. Hanabi is trained with three seeds, and Dou Dizhu, Guan Dan and Dark Chess with one. Dark Chess starts with 5,000 iterations of imitating a chess policy that sees the full board and uses no test-time policy improvement, and its 40,000 iterations count the self-play that follows.

\begin{table}[ht]
\centering\small
\setlength{\tabcolsep}{4pt}
\begin{tabular}{@{}lcccccccc@{}}
\toprule
Game & $d$ & $W$ & $\eta$ & $T$ & Environments & Steps & Iterations & Transitions\\
\midrule
Hanabi 2P & 192 & 32 & $4\times10^{-4}$ & 10,000 & 512 & 512 & 40,000 & $1.05\times10^{10}$\\
Hanabi 3P & 192 & 32 & $4\times10^{-4}$ & 10,000 & 512 & 512 & 40,000 & $1.05\times10^{10}$\\
Hanabi 4P & 192 & 32 & $4\times10^{-4}$ & 10,000 & 512 & 512 & 40,000 & $1.05\times10^{10}$\\
Hanabi 5P & 192 & 32 & $4\times10^{-4}$ & 10,000 & 512 & 512 & 40,000 & $1.05\times10^{10}$\\
\midrule
Dou Dizhu & 384 & 64 & $2\times10^{-4}$ & 10,000 & 1,024 & 512 & 40,000 & $2.10\times10^{10}$\\
Guan Dan & 384 & 64 & $2\times10^{-4}$ & 10,000 & 1,024 & 512 & 40,000 & $2.10\times10^{10}$\\
Dark Chess & 384 & 64 & $2\times10^{-4}$ & 10,000 & 1,024 & 512 & 40,000 & $2.10\times10^{10}$\\
\bottomrule
\end{tabular}

\caption{Hyperparameters by game: the residual width $d$ and history window $W$ of the actor, the base learning rate $\eta$, the schedule constant $T$, the parallel environments, the steps per environment and iteration, the iterations and the environment transitions of a run. In a transition every player receives an observation. Dark Chess counts self-play after its imitation phase.}
\label{tab:hyperparameters}
\end{table}

At iteration $i$, the actor learning rate, the critic learning rate and the clipping coefficient are
\begin{equation}
 \eta_{\mathrm{actor}}=\eta\,f(i),\qquad
 \eta_{\mathrm{critic}}=\eta\,\sqrt{f(i)},\qquad
 \eps=\eps_{\mathrm{base}}\,f(i),\qquad
 f(i)=\min\Bigl\{1,\frac{T}{i}\Bigr\}.
\end{equation}
Every evaluation uses the exponential moving average of the actor's weights with decay $0.999$.

\paragraph{Model Architecture.}
The actor is the \FTT\ of \cref{sec:method} with $L=6$ blocks. Both attentions have four heads and width $2d/3$, the feed-forward networks have width $2d$, and the residual width $d$ and the history window $W$ of each game are given in \cref{tab:hyperparameters}. The centralized critic has its own parameters and is used only in training; it uses $L=6$ blocks with field attention only over the fields of the full state, with the same feature and FFN widths as the corresponding \FTT\ actor.

\clearpage
\section{Evaluation Details}
\label{app:evaluation}

\subsection{Hanabi}
We evaluate Hanabi policies in self-play with the averaged weights of Appendix~\ref{app:training}, taking the most probable action at every decision. A game that loses all its lives scores zero. Checkpoints are saved every 1,000 iterations and evaluated on 2,000 games, the final checkpoint on 5,000, and all checkpoints and actors share the same random seeds. All 30 training runs use the hyperparameters of \cref{tab:hyperparameters}. We checked our environment against the Hanabi Learning Environment~\citep{bard2019hanabi} by replaying 40 games with 2,478 decisions, and the two agreed in every observation and outcome.

\paragraph{Actor Compute.}
The FLOPs of \cref{tab:hanabi-budgets} are the floating-point additions and multiplications of one decision by one player's actor with a full history window, including the update of its memory. A multiply-add counts as two operations. The count covers the embedding, the blocks and the readout, and it leaves out the environment, the critic and training.

\paragraph{Published Reference Values.}
The \FTT\ rows of \cref{tab:hanabi-published} are means over the three training seeds, with no selection among the runs. The reference rows are taken from the publications, and each keeps the training and evaluation budget of its paper. The SAD values are from Table~2 of \citet{hu2019sad}, where every entry is the best of 13 training runs, evaluated on 100,000 games with a score of zero when all lives are lost. For SPARTA we take, at each player count, the highest-scoring variant in Tables~1 and~2 of \citet{lerer2019sparta}: multi-agent search with a maximum range of 2,000 for two players and single-agent search for three to five players, all on learned blueprints. Both papers report standard errors for the scores and none for the perfect rates of these policies, so those cells of the table are blank. The values of \citet{forkel2025entropy} are the greedy self-play scores of IPPO with an LSTM and entropy coefficient $0.01$ in their Table~1, the variant with the highest self-play score at every player count; each is a mean over four training seeds, five in five-player Hanabi, each evaluated on 5,000 games, with the standard error over the seeds, and the paper reports no perfect rates. The Ataraxos values are from Supplementary Table~S32 of \citet{sokota2026ataraxos}, with 10,000 evaluation games and, for test-time policy improvement, 10,000 rollouts per decision by every player. Values that a publication gives with fewer than three decimals are padded with zeros.

\subsection{Dou Dizhu and Guan Dan}
\Cref{sec:competitive} defines the gain and the exchange of roles. For Dou Dizhu deal $j$, let $r^L_j$ be the landlord's score when the candidate is the landlord and $r^P_j$ the landlord's score when the reference is the landlord; the gain on that deal is $g_j=(r^L_j-r^P_j)/2$. The benchmark has no bidding, so the actor's history starts with a fixed bidding sequence that selects the benchmark's landlord without doubling. In Guan Dan our rules engine is the referee, DanZero plays with its released weights and its own observation encoder, the opening seat is fixed, and each deal is played under both team assignments.

The final \FTT\ policy plays 10,000 paired deals each against PerfectDou, DouZero and Ataraxos without test-time policy improvement, 10,000 against Ataraxos with test-time policy improvement, and 2,000 against DanZero. A paired deal is two hands with the roles exchanged, and the 10,000 deals against DouZero and against Ataraxos are all distinct. DouZero uses its released ADP weights for each of its three roles, and all \FTT\ checkpoints use averaged weights. Ataraxos faces the \FTT\ agent on the same 10,000 deals and the same role pairing with and without test-time policy improvement, so the two results can be compared deal by deal: test-time policy improvement recovers $0.042\pm0.018$ of the $0.236$ points per deal.

For Ataraxos we use the 40,000-iteration policy checkpoint of its original Dou Dizhu experiments, which takes the most probable action when it plays without test-time policy improvement. With test-time policy improvement, the same policy is combined with its 40,000-iteration belief checkpoint and improved by an update-equivalent step with terminal-rollout action values and a step size of 5; the belief-sampling temperature is 1, and the rollout and final-action temperatures are 0. The nominal rollout budget is 200 per decision: each searched action receives $\lceil200/A\rceil$ belief-world rollouts, where $A$ is the number of searched actions, so rounding can make the total exceed 200. To check our Ataraxos adapter, we replayed the published comparison of its base policy against PerfectDou on 1,000 separate pairs and obtained $0.159\pm0.049$ points, consistent with the published $0.152\pm0.004$.

In \cref{fig:cards}, the horizontal lines and the 20,000-iteration points come from matches in which the final policy is the candidate. We plot them from the other side, with every paired gain negated and its standard error unchanged. The curves start from a uniform-random policy, which samples uniformly among the legal plays, including a pass when it is legal, over 1,000 paired deals in Dou Dizhu and 500 in Guan Dan.

\subsection{Dark Chess}
The round robin of \cref{sec:competitive} plays every pair of the 21 checkpoints at iterations 0, 2,000 and so on up to 40,000, where iteration 0 is the policy at the start of self-play, after the imitation phase. A pairing consists of 256 pairs of games, each pair played with the colors exchanged and the same random seed. Both policies use averaged weights, sample their first five moves and then take the most probable action. A game counts one for a win, one half for a draw and zero for a loss, and the standard error of a pairing is taken over its 256 pair means.

We fit one rating per checkpoint by maximum likelihood under the Elo model, in which a checkpoint rated $r_i$ is expected to score $1/(1+10^{(r_j-r_i)/400})$ against one rated $r_j$ and a draw counts as half a win. The rating of the first checkpoint is set to zero, and \cref{fig:darkchess-curves} shows every rating relative to the final checkpoint. There, each of the tree sizes 1,000, 10,000, 100,000 and 1,000,000 of the reimplemented search is placed at the rating that maximizes the likelihood of its matches against the checkpoints after 10,000, 20,000, 30,000 and 40,000 iterations under the same model, with the ratings of the checkpoints held fixed; a match with score $s$ and standard error $e$ over its pairs counts as $s(1-s)/e^2$ games. The fit underestimates the score of the checkpoint after 10,000 iterations against every tree size, by $2.7$ to $4.8$ standard errors, and overestimates that of the final checkpoint, by $1.6$ to $3.4$, so the round robin spreads the checkpoints farther apart than their matches against the search do. The standard errors of the ratings, about 7 points, come from resampling the 256 pairs of every pairing. The ratings are fitted to all 210 pairings at once, so a rating measures the strength of a checkpoint against the whole set of checkpoints.

\paragraph{Reimplementation of the Search of Obscuro.}
The opponent of \cref{sec:competitive} is our C++ reimplementation of the search of \citet{zhang2026obscuro}, whose authors have released no code. It follows the paper in the components of the search: the set of possible positions is kept exactly, the subgame is the 2-KLUSS subgame over the tree kept from the previous move, it is solved by PCFR+ with the Resolve and Maxmargin gadgets, the tree is expanded by one-sided GT-CFR, and leaves are valued by Stockfish 14 at depth 1. It differs from the paper in four points. The budget is the number of nodes of the search tree at the moment the move is chosen, in place of a time limit; the paper reports 1,070,552 nodes on average at 5 seconds per move. One thread interleaves the paper's solver thread and its two expansion threads in a fixed order. The rules are those of our environment (Appendix~\ref{app:games}), in which a move that leaves the mover's king attacked ends the game. Repetition and the fifty-move rule are not modelled inside the tree. Two choices that the paper leaves open were set so that the first moves of the engine match the opening statistics of the paper. As a check, the engine played itself at 1,000,000 nodes. White scores $0.570$, against $0.575$ in the paper, and the tree has 999,639 nodes and 12,858 information sets on average, against 1,070,552 and 14,404. The games have more draws than those of the paper ($0.257$ against $0.162$), are shorter (105.1 plies against 116.6) and have larger sets of possible positions (135,055 on average against 17,264). Our results are results against this reimplementation.

\paragraph{Matches against the Reimplementation.}
The final policy plays 256 pairs of games with the colors exchanged against each of the eight tree sizes 1,000, 3,000, 10,000, 30,000, 100,000, 300,000, 1,000,000 and 3,000,000, with the same seed for a pair at every size, and the policies after 10,000, 20,000 and 30,000 iterations play 128 pairs each against the trees of 1,000, 10,000, 100,000 and 1,000,000 nodes. The \FTT\ agent plays as in the round robin, with averaged weights, its first five moves sampled and the most probable action afterwards. It runs on CPU in these matches and chooses the move of the round-robin harness in $98.8\%$ of its decisions after 40,000 iterations and in $99.1\%$ to $99.4\%$ after 10,000, 20,000 and 30,000. The score is the mean over the complete pairs, and the standard error is taken over the pair means. Of the 7,168 games, 51 have no result, 49 because the engine's set of possible positions passed $10^9$ and two because the engine gave no answer within two hours. These games are not scored, and a pair counts only when both of its games do; counting all of them as lost or as won by the \FTT\ agent changes no score by more than $1.3$ points.

\clearpage
\section{More Experiment Results}
\label{app:results}

\subsection{Hanabi}
\Cref{tab:hanabi-seeds} gives the final result of every training run behind \cref{fig:hanabi,tab:hanabi-published}.

\begin{table}[!ht]
\centering\small
\begin{tabular}{@{}llcr@{}lr@{}l@{}}
\toprule
Game & Agent & Seed & \multicolumn{2}{c}{Score} & \multicolumn{2}{c}{Perfect rate (\%)}\\
\midrule
Hanabi 2P & FTT & 0 & $24.771$ & ${}\pm0.009$ & $84.400$ & ${}\pm0.513$\\
 &  & 1 & $24.773$ & ${}\pm0.009$ & $84.540$ & ${}\pm0.511$\\
 &  & 2 & $24.788$ & ${}\pm0.008$ & $85.300$ & ${}\pm0.501$\\
\addlinespace[2pt]
 & FTT, field-attention only & 0 & $24.707$ & ${}\pm0.010$ & $81.360$ & ${}\pm0.551$\\
 &  & 1 & $24.743$ & ${}\pm0.009$ & $82.860$ & ${}\pm0.533$\\
 &  & 2 & $24.744$ & ${}\pm0.009$ & $83.020$ & ${}\pm0.531$\\
\addlinespace[2pt]
 & LSTM, compute-matched & 0 & $24.552$ & ${}\pm0.012$ & $72.700$ & ${}\pm0.630$\\
 &  & 1 & $24.519$ & ${}\pm0.015$ & $71.620$ & ${}\pm0.638$\\
 &  & 2 & $24.579$ & ${}\pm0.014$ & $75.180$ & ${}\pm0.611$\\
\addlinespace[2pt]
 & LSTM, parameter-matched & 0 & $24.485$ & ${}\pm0.014$ & $68.840$ & ${}\pm0.655$\\
 &  & 1 & $24.464$ & ${}\pm0.014$ & $69.000$ & ${}\pm0.654$\\
 &  & 2 & $24.452$ & ${}\pm0.015$ & $68.540$ & ${}\pm0.657$\\
\midrule
Hanabi 3P & FTT & 0 & $24.964$ & ${}\pm0.003$ & $96.980$ & ${}\pm0.242$\\
 &  & 1 & $24.959$ & ${}\pm0.006$ & $97.340$ & ${}\pm0.228$\\
 &  & 2 & $24.964$ & ${}\pm0.003$ & $97.040$ & ${}\pm0.240$\\
\addlinespace[2pt]
 & FTT, field-attention only & 0 & $24.953$ & ${}\pm0.004$ & $96.340$ & ${}\pm0.266$\\
 &  & 1 & $24.944$ & ${}\pm0.006$ & $96.140$ & ${}\pm0.272$\\
 &  & 2 & $24.948$ & ${}\pm0.004$ & $95.820$ & ${}\pm0.283$\\
\addlinespace[2pt]
 & LSTM, compute-matched & 0 & $24.728$ & ${}\pm0.013$ & $82.840$ & ${}\pm0.533$\\
 &  & 1 & $24.741$ & ${}\pm0.013$ & $84.080$ & ${}\pm0.517$\\
 &  & 2 & $24.737$ & ${}\pm0.011$ & $83.280$ & ${}\pm0.528$\\
\addlinespace[2pt]
 & LSTM, parameter-matched & 0 & $24.605$ & ${}\pm0.015$ & $76.460$ & ${}\pm0.600$\\
 &  & 1 & $24.648$ & ${}\pm0.011$ & $78.060$ & ${}\pm0.585$\\
 &  & 2 & $24.555$ & ${}\pm0.013$ & $73.440$ & ${}\pm0.625$\\
\midrule
Hanabi 4P & FTT & 0 & $24.949$ & ${}\pm0.004$ & $95.800$ & ${}\pm0.284$\\
 &  & 1 & $24.951$ & ${}\pm0.004$ & $96.000$ & ${}\pm0.277$\\
 &  & 2 & $24.946$ & ${}\pm0.004$ & $95.720$ & ${}\pm0.286$\\
\midrule
Hanabi 5P & FTT & 0 & $24.869$ & ${}\pm0.006$ & $89.360$ & ${}\pm0.436$\\
 &  & 1 & $24.866$ & ${}\pm0.006$ & $88.860$ & ${}\pm0.445$\\
 &  & 2 & $24.855$ & ${}\pm0.006$ & $88.180$ & ${}\pm0.457$\\
\bottomrule
\end{tabular}

\caption{Hanabi score and perfect rate after 40,000 iterations for each training seed, mean $\pm1$ SEM over 5,000 evaluation games. These within-run errors differ from the across-seed SEMs in \cref{fig:hanabi,tab:hanabi-published}.}
\label{tab:hanabi-seeds}
\end{table}

\subsection{Dark Chess}
\label{app:darkchess}
\begin{figure}[!ht]
\centering
\includegraphics[width=0.40\linewidth]{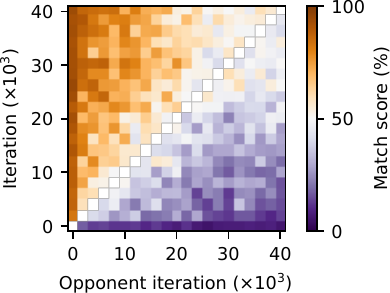}
\caption{Dark Chess round robin: the match score of the checkpoint on the vertical axis against the checkpoint on the horizontal axis, over 512 games per pairing.}
\label{fig:darkchess}
\end{figure}

The round robin behind the Dark Chess ratings of \cref{fig:darkchess-curves} has 210 pairings and 107,520 games, of which $7\%$ are draws; Appendix~\ref{app:evaluation} gives its protocol. A later checkpoint scores $55\%$ on average against the checkpoint 2,000 iterations before it, $65\%$ at a distance of 10,000 iterations and $74\%$ at 20,000. Play among the checkpoints is not fully transitive, and \cref{fig:darkchess} gives the score of every single pairing.

\Cref{tab:darkchess-obscuro} lists the matches against our reimplementation of the search of Obscuro (Appendix~\ref{app:evaluation}). The final policy scores above one half at every tree size, and against the tree of 1,000,000 nodes the score rises with training and reaches one half after 20,000 iterations.

\begin{table}[!ht]
\centering\small
\begin{tabular}{@{}rrrrrrr@{}l@{}}
\toprule
Tree nodes & \FTT\ iterations & Pairs & Wins & Draws & Losses & \multicolumn{2}{c}{Score (\%)}\\
\midrule
1,000 & 40,000 & 254 & 441 & 6 & 61 & $87.4$ & ${}\pm1.5$\\
3,000 & 40,000 & 252 & 422 & 8 & 74 & $84.5$ & ${}\pm1.7$\\
10,000 & 40,000 & 252 & 392 & 15 & 97 & $79.3$ & ${}\pm1.8$\\
30,000 & 40,000 & 252 & 360 & 25 & 119 & $73.9$ & ${}\pm1.9$\\
100,000 & 40,000 & 253 & 313 & 30 & 163 & $64.8$ & ${}\pm2.1$\\
300,000 & 40,000 & 252 & 278 & 40 & 186 & $59.1$ & ${}\pm2.1$\\
1,000,000 & 40,000 & 250 & 276 & 44 & 180 & $59.6$ & ${}\pm2.0$\\
3,000,000 & 40,000 & 253 & 252 & 60 & 194 & $55.7$ & ${}\pm2.1$\\
\midrule
1,000 & 10,000 & 126 & 197 & 3 & 52 & $78.8$ & ${}\pm2.7$\\
1,000 & 20,000 & 126 & 206 & 5 & 41 & $82.7$ & ${}\pm2.3$\\
1,000 & 30,000 & 128 & 218 & 5 & 33 & $86.1$ & ${}\pm2.2$\\
10,000 & 10,000 & 127 & 170 & 10 & 74 & $68.9$ & ${}\pm2.9$\\
10,000 & 20,000 & 126 & 189 & 9 & 54 & $76.8$ & ${}\pm2.5$\\
10,000 & 30,000 & 124 & 191 & 9 & 48 & $78.8$ & ${}\pm2.4$\\
100,000 & 10,000 & 127 & 130 & 23 & 101 & $55.7$ & ${}\pm2.8$\\
100,000 & 20,000 & 125 & 141 & 20 & 89 & $60.4$ & ${}\pm2.9$\\
100,000 & 30,000 & 127 & 153 & 24 & 77 & $65.0$ & ${}\pm2.8$\\
1,000,000 & 10,000 & 127 & 96 & 32 & 126 & $44.1$ & ${}\pm3.0$\\
1,000,000 & 20,000 & 125 & 114 & 24 & 112 & $50.4$ & ${}\pm3.2$\\
1,000,000 & 30,000 & 127 & 143 & 12 & 99 & $58.7$ & ${}\pm3.1$\\
\bottomrule
\end{tabular}

\caption{Dark Chess: the \FTT\ agent against our reimplementation of the search of Obscuro. Wins, draws and losses are those of the \FTT\ agent over the complete pairs of games with the colors exchanged. The score is the mean over the pairs with $\pm1$ SEM.}
\label{tab:darkchess-obscuro}
\end{table}

\end{document}